\documentclass[conference, 10pt]{IEEEtran}
\IEEEoverridecommandlockouts
\usepackage{cite}
\usepackage{amsmath,amssymb,amsfonts}
\usepackage{graphicx}
\usepackage{textcomp}
\usepackage{xcolor}
\usepackage{ dsfont }
\usepackage{booktabs}
\usepackage{tablefootnote}
\usepackage{pbox}
\usepackage{enumitem}
\usepackage{pifont}
\usepackage{tikz}
\usepackage{inconsolata}
\usepackage{algorithm}
\usepackage{algorithmicx}
\usepackage{lipsum}
\newcommand{\RomanNumeralCaps}[1]
    {\MakeUppercase{\romannumeral #1}}
\usepackage[noend]{algpseudocode}

\algnewcommand\algorithmicforeach{\textbf{for each}}
\algdef{S}[FOR]{ForEach}[1]{\algorithmicforeach\ #1\ \algorithmicdo}

\algnewcommand{\IIf}[1]{\State\algorithmicif\ #1\  \algorithmicthen}
\algnewcommand{\EndIIf}{\unskip}
\usepackage{mathrsfs}

\usepackage{multirow}
\usepackage{longtable}

\usepackage[colorlinks,linkcolor = blue, urlcolor  = blue, citecolor = blue, anchorcolor = green ,bookmarks=false,hypertexnames=true]{hyperref}

\def\BibTeX{{\rm B\kern-.05em{\sc i\kern-.025em b}\kern-.08em
    T\kern-.1667em\lower.7ex\hbox{E}\kern-.125emX}}

\newcommand\encircle[1]{%
  \tikz[baseline=(X.base)] 
    \node (X) [draw, shape=circle, inner sep=-1, fill=black, text=white] {\strut #1};%
}

\usepackage{xspace}
\newcommand{\Design}{\textit{$\mathsf{DOMINO}$\xspace}}    

\begin{document}


\title{DOMINO: \underline{Dom}ain-\underline{I}nvariant Hyperdime\underline{n}si\underline{o}nal Classification for Multi-Sensor Time Series Data\vspace{-8mm}}

\author{
    \IEEEauthorblockN{Junyao Wang$^\dag$, Luke Chen$^\S$, Mohammad Abdullah Al Faruque$^\dag$$^\S$}
    \IEEEauthorblockN{\textit{$^\dag$ Department of Computer Science, University of California, Irvine, CA, United States}\\
    $^\S$ \textit{Department of Electrical Engineering and Computer Science, University of California, Irvine, CA, United States}
    \\\textit{\{junyaow4, panwangc, alfaruqu\}}@uci.edu\vspace{-7mm}}
}

\maketitle

\begin{abstract}
With the rapid evolution of the Internet of Things, many real-world applications utilize heterogeneously connected sensors to capture time-series information. 
Edge-based machine learning (ML) methodologies are often employed to analyze locally collected data. 
However, a fundamental issue across data-driven ML approaches is \textit{distribution shift}. 
It occurs when a model is deployed on a data distribution different from what it was trained on, and can substantially degrade model performance. 
Additionally, increasingly sophisticated deep neural networks (DNNs) have been proposed to capture spatial and temporal dependencies in multi-sensor time series data, requiring intensive computational resources beyond the capacity of today's edge devices. 
While brain-inspired hyperdimensional computing (HDC) has been introduced as a lightweight solution for edge-based learning, existing HDCs are also vulnerable to the distribution shift challenge. 
In this paper, we propose $\Design$, a novel HDC learning framework addressing the distribution shift problem in noisy multi-sensor time-series data. 
$\Design$ leverages efficient and parallel matrix operations on high-dimensional space to dynamically identify and filter out domain-variant dimensions.  
Our evaluation on a wide range of multi-sensor time series classification tasks shows that $\Design$ achieves on average $2.04\%$ higher accuracy than state-of-the-art (SOTA) DNN-based domain generalization techniques, and delivers $16.34\times$ faster training and $2.89\times$ faster inference.
More importantly, $\Design$ exhibits notably better performance when learning from partially labeled data and highly imbalanced data, and provides $10.93\times$ higher robustness against hardware noises than SOTA DNNs. 

\end{abstract}


\section{Introduction}\label{sec:intro}
The Internet of Things (IoT) has become an emerging trend for its extraordinary potential to connect heterogeneous devices and enable them with new capabilities~\cite{wang2023disthd}.
Many real-world IoT applications utilize multiple sensors to collect information over the course of time, constituting multi-sensor time series data~\cite{shrivastwa2020brain, rashid2022template, demirel2022neural}. 
These applications often leverage edge-based machine learning (ML) algorithms to analyze locally collected data and perform various learning tasks. 
However, a critical issue across data-driven ML approaches, including deep neural networks (DNNs), is \textit{distribution shift}. 
In particular, the excellent performance of these ML algorithms relies heavily on the critical assumption that the training and inference data are independently and identically distributed (i.i.d.); i.e., they come from the same distribution~\cite{dong2022first}.
Unfortunately, this assumption can be easily violated in real-world scenarios and have shown to substantially degrade model performance in many embedded ML applications, where instances from unseen domains not fitting the distribution of the training data are inevitable~\cite{pooch2020can, subbaswamy2020development, wilson2023hyperdimensional}. 
For instance, in the field of mobile health, models can systematically fail when tested on patients from different hospitals or people from diverse demographics~\cite{gulrajani2021in}. 

A number of innovative \textit{domain generalization} (DG) techniques have been proposed for deep learning (DL)~\cite{wang2018deep, zhao2020review}. 
However, due to their weak notion of memorization, these DL approaches often fail to 
perform well on noisy multi-sensor time series data with spatial and temporal dependencies~\cite{wilson2020multi}. 
Recurrent neural networks (RNNs), e.g., long short-term memory (LSTM), have recently been proposed to address this issue~\cite{qin2017dual, su2019robust, hochreiter1997long}. 
Nevertheless, these models are notably complicated and inefficient to train, and their intricate architectures require substantial off-chip memory and computational power to 
iteratively refine millions of parameters over multiple time periods~\cite{hochreiter1997long}. 
Such resource-intensive requirements can be impractical for less powerful computing platforms. 
Considering the massive amount of information nowadays, the power and memory limitations of embedded devices, and the potential instabilities of IoT systems, a more lightweight, efficient, and scalable learning framework to combat the distribution shift issue in multi-sensor time series data are of critical need.   

\begin{figure}[!t]
\centering
\includegraphics[width=\linewidth]{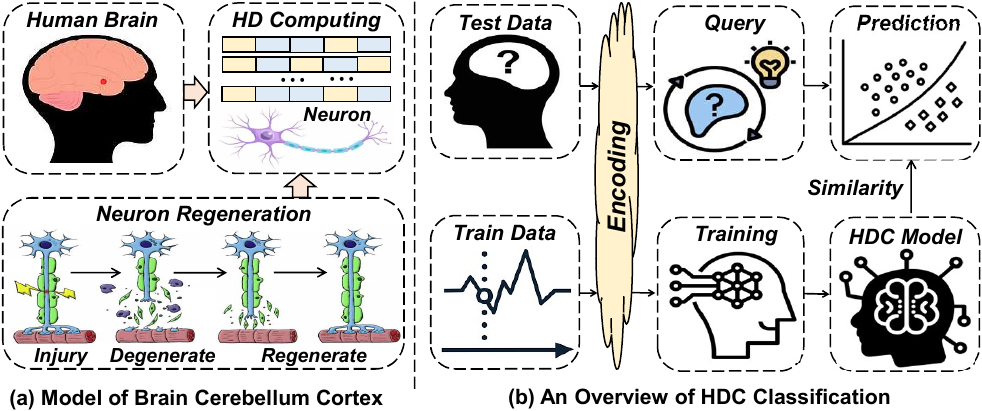}
\vspace{-7mm}
\caption{An Overview of Brain Cerebellum Cortex and HDC Classification}
\vspace{-6.5mm}
\label{fig: intro}  
\end{figure}


In contrast to traditional AI methodologies, brain-inspired Hyperdimensional Computing (HDC) incorporates learning capability along with typical memory functions of storing/loading information, and hence brings unique advantages in dealing with time-series data~\cite{ni2022neurally}. 
Additionally, HDC provides a powerful learning solution for today's edge platforms by providing notably fast convergence, high computational efficiency, ultra-robustness against noises, and lightweight hardware implementation~\cite{imani2019framework, ge2020classification, zou2021scalable}. 
As demonstrated in Fig. \ref{fig: intro}(a), HDC is motivated by the neuroscience observation that the cerebellum cortex in human brains 
effortlessly and efficiently process memory, perception, and cognition information without much concern for noisy or broken neuron cells. 
Closely mimicking the information representation and functionalities of human brains, HDC \textit{encodes} low-dimensional inputs to \textit{hypervectors} with thousands of elements to perform various learning tasks~\cite{zou2021scalable} as shown in Fig. \ref{fig: intro}(b). 
HDC then conducts highly parallel and well-trackable operations, and has been proven to achieve high-quality results in various learning tasks with comparable accuracy to SOTA DNNs~\cite{ge2020classification, zou2021scalable, wang2023late}. 
\begin{figure}[!t]
\centering
\includegraphics[width=\linewidth]{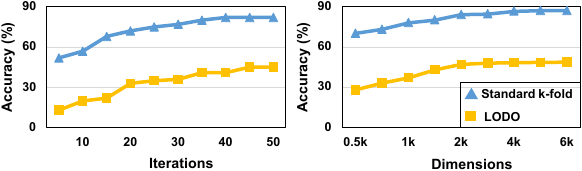}
\vspace{-7mm}
\caption{Comparison of the Accuracy of LODO CV and Standard $k$-fold CV}
\vspace{-5mm}
\label{fig: motiv}  
\end{figure}

Unfortunately, existing HDCs are not immune to the distribution shift issue. 
As shown in Fig. \ref{fig: motiv}, SOTA HDCs converge at notably lower accuracy in leave-one-domain-out (LODO) cross-validation (CV) than in standard $k$-fold CV regardless of training iteration and model complexity.
LODO  CV involves training a model on all the available data except for one domain that is left out for inference, while standard $k$-fold CV randomly divides all data into $k$ subsets with $k-1$ subsets for training and the remaining one for inference. 
Such performance degradation indicates a very limited generalization capability of existing HDCs. 
However, standard $k$-fold CV does not reflect the real-world distribution shift problem, as the random sampling process introduces \textit{data leakage} that enables the training data to include information from all the domains. 
The accuracy from the standard $k$-fold CV is thus often considerably higher than that in real-world scenarios.
In contrast, with constant dynamic regeneration of cerebral neurons shown in Fig. \ref{fig: intro}(a), humans are capable of effectively identifying and recalling attributes shared across different domains, and naturally filtering out information that is too specific and biased towards a single domain. 
Humans then utilize this shared knowledge to infer instances from novel domains. 
While the goal of HDC is to exploit the high-dimensionality of randomly generated vectors 
to represent information mimicking a pattern of neural activity, it remains challenging for existing HDCs to support a similar behavior. 

In this paper, we propose $\Design$, a novel HDC domain generalization (DG) algorithm for multi-sensor time series classification. $\Design$ effectively eliminates dimensions representing domain-variant information to enhance model generalization capability. Our main contributions are listed below: 
\begin{itemize}[leftmargin = *]
\item To the best of our knowledge, $\Design$ is the first HDC-based DG algorithm. 
By dynamically identifying and regenerating domain-variant dimensions, $\Design$ achieves on average $2.04\%$ higher accuracy than SOTA DL-based techniques with $16.34\times$ faster training and $2.89\times$ faster inference, ensuring accurate and timely performance for DG.
\item 
$\Design$ achieves up to $5.81\%$ higher accuracy than SOTA DNNs in scenarios with a significantly limited amount of labeled data and on average $2.58\%$ higher accuracy when learning from highly imbalanced data across different domains. Additionally, $\Design$ exhibits $10.93\times$ higher robustness against hardware noise than SOTA DNNs. 
\item We propose hardware-aware optimizations for $\Design$'s implementation, and evaluated it across multiple embedded hardware devices including Raspberry Pi and NVIDIA Jetson Nano. $\Design$ exhibits considerably lower inference latency and energy consumption than DL-based approaches. 
\end{itemize}


\begin{figure*}[t]
  \centering
  {\includegraphics[width=\textwidth]{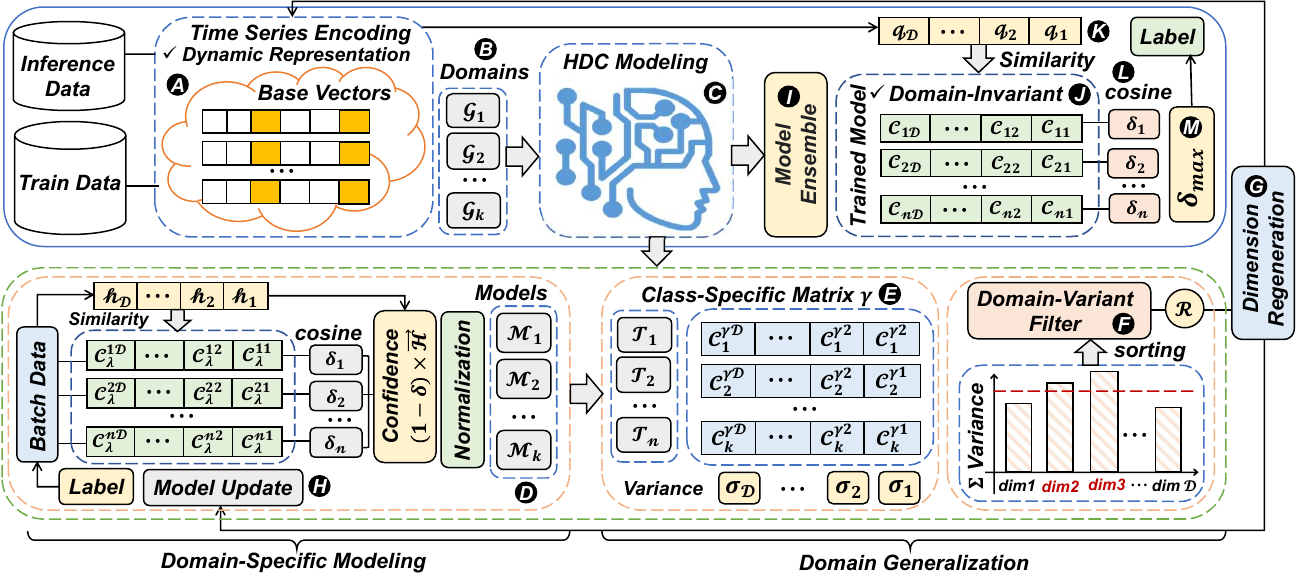}}
  \vspace{-6mm}
  \caption{An Overview of $\Design$ Workflow }
 \vspace{-2mm}
  \label{fig: flow} 
\end{figure*}

\section{Methodology} \label{sec:method}
The goal of $\Design$ is to effectively leverage the information learned from each training iteration to identify and filter out domain-variant dimensions, thereby enhancing the generalization capability of our model. 
As shown in Fig. \ref{fig: flow}, $\Design$ starts with encoding multi-sensor time series data samples into high-dimensional space $(\encircle{A})$. 
$\Design$ then conducts two innovative steps, \textit{domain-specific modeling} and \textit{domain generalization}, to enable its encoding module and randomly-generated base vectors with awareness of the relevance of each dimension to the domains. 
In each iteration of \textit{domain-specific modeling}, $\Design$ exploits an efficient and parallel hyperdimensional learning algorithm to construct domain-specific hyperdimensional models $(\encircle{D})$. 
In \textit{domain generalization}, we utilize domain-specific models to form class-specific matrices $(\encircle{E})$, and then identify and filter out dimensions that represent domain-variant information$(\encircle{F})$. 
To mitigate the performance degradation caused by eliminating dimensions, we replace these dimensions with randomly generated hypervectors and retrain them. 
Note that all these operations can be done in a highly parallel matrix-wise way, as multiple training samples can be grouped into a matrix of row hypervectors.

\subsection{HDC Preliminaries}  \label{sec:pre}
Inspired by the high-dimensional information representation 
in human brains, 
HDC maps inputs onto hyperdimensional space as \textit{hypervectors} $(\encircle{A})$, each of which contains thousands of elements. 
One unique property of the hyperdimensional space is the existence of a large number of nearly orthogonal hypervectors, enabling highly parallel operations such as similarity calculations, bundlings, and bindings. 
Mathematically, consider random bipolar hypervectors ${\mathcal{H}}_1$ and ${\mathcal{H}}_2$ with dimension $\mathcal D$, i.e., ${\mathcal{H}}_1, {\mathcal{H}}_2 \in \{-1,1\}^{\mathcal D}$, when $\mathcal D$ is large enough, the dot product ${\mathcal{H}}_1\cdot  {\mathcal{H}}_2\approx 0$. 
\textbf{Similarity:} calculation of the distance between the query hypervector and the class hypervector (noted as $\delta(\cdot, \cdot)$). For real-valued hypervectors,  a common measure is cosine similarity. 
For bipolar hypervectors, it is simplified to the Hamming distance. 
\textbf{Bundling (+):} element-wise addition of multiple hypervectors, e.g., ${\mathcal{H}}_{bundle}= {\mathcal{H}}_1 +{\mathcal{H}}_2$, generating a hypervector with the same dimension as inputs. In high-dimensional space, bundling works as a memory operation and provides an easy way to check the existence of a query hypervector in a bundled set. In the previous example,  $\delta({\mathcal{H}}_{bundle},{\mathcal{H}}_1)\gg 0 $ while $\delta({\mathcal{H}}_{bundle},{\mathcal{H}}_3)\approx 0 $ (${\mathcal{H}}_3\neq {\mathcal{H}}_1, {\mathcal{H}}_2)$. 
\textbf{Binding (*):} element-wise multiplication associating two hypervectors to create another near-orthogonal hypervector, i.e. ${\mathcal{H}}_{bind} ={\mathcal{H}}_{1}* {\mathcal{H}}_{2}$ 
 where $\delta({\mathcal{H}}_{bind}, {\mathcal{H}}_{1})\approx 0$ and $\delta({\mathcal{H}}_{bind}, {\mathcal{H}}_{2})\approx 0$. 
 Due to reversibility, in bipolar cases,
 ${\mathcal{H}}_{bind} * {\mathcal{H}}_1 = {\mathcal{H}}_2$, information from both hypervectors can be preserved. 
 Binding models how human brains \textit{connect} input information. 
  \textbf{Permutation $(\rho)$}: a single circular shift of a hypervector by moving the value of the final dimension to the first dimension and shifting all other values to their next dimension. It generates a permuted hypervector that is nearly orthogonal to its original hypervector, i.e., $\delta(\rho \mathcal H, \mathcal H)\approx 0$. Permutation models how human brains handle \textit{sequential} information.
 \textbf{Inference:} The inference of HDC consists of two steps: (\romannumeral 1) encode $(\encircle{A})$ inference data with the same encoder in training to generate a query hypervector $ {\mathcal{Q}}$ $(\encircle{K})$, and (\romannumeral 2) calculate the distance or similarity between $ {\mathcal{Q}}$ and each class hypervector $(\encircle{L})$. 
 $ {\mathcal{Q}}$ is then classified to the class where it achieves the highest similarity$(\encircle{M})$. 

\subsection{Encoding of Multi-Sensor Time Series} \label{sec:encode}
\begin{figure}[!t]
\centering
\includegraphics[width=\linewidth]{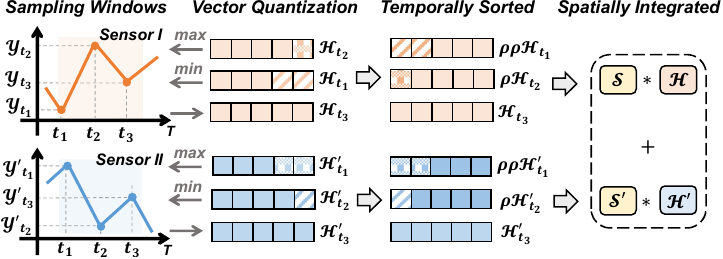}
\vspace{-4mm}
\caption{HDC Encoding Technique for Multi-Sensor Time Series Data}
\vspace{-4mm}
\label{fig: tsencode}  
\end{figure}
To capture the temporal and spatial dependencies in multi-sensor time series data, we employ the encoding techniques demonstrated in Fig. \ref{fig: tsencode}.  
We sample time series data in $n$-gram windows; in each sample window, the signal values ($y$-axis) store the information and the time ($x$-axis) represents the temporal sequence. 
We first assign random hypervectors $\mathcal H_{max}$ and $\mathcal H_{min}$ to represent the maximum and minimum signal values, i.e., $y_{max}$ and $y_{min}$. 
We then perform vector quantization to values between $y_{max}$ and $y_{min}$ to generate vectors that have a spectrum of similarity to $\mathcal H_{max}$ and $\mathcal H_{min}.$ 
As shown in Fig. \ref{fig: tsencode}, Sensor \RomanNumeralCaps{1} and Sensor \RomanNumeralCaps{2} follow a time series in trigram. 
Sensor \RomanNumeralCaps{1} has the maximum value at $t_2$ and the minimum value at $t_1$, and hence we assign randomly generated hypervectors $\mathcal H_{t_2}$ and $\mathcal H_{t_1}$ to $y_{t_2}$ and $y_{t_1}$. Similarly, we assign randomly generated hypervectors $\mathcal H'_{t_1}$ and $\mathcal H'_{t_2}$ to $y'_{t_1}$ and $y'_{t_2}$ in Sensor \RomanNumeralCaps{2}. We then assign hypervectors to $y_{t_3}$ in Sensor \RomanNumeralCaps{1} and value at $y'_{t_3}$ in Sensor \RomanNumeralCaps{2} through vector quantization, i.e., 
\begin{align*}
    &\mathcal H_{t_3} = \mathcal H_{t_1}+\frac{y_{t_3}-y_{t_1}}{y_{t_2}-y_{t_1}}\cdot (\mathcal H_{t_2} - \mathcal H_{t_1})\\
    & \mathcal H'_{t_3} = \mathcal H'_{t_2}+\frac{y'_{t_3}-y'_{t_2}}{y'_{t_1}-y'_{t_2}}\cdot (\mathcal H'_{t_1} - \mathcal H'_{t_2}).
\end{align*}
We then represent the temporal sequence of data samples utilizing permutations as explained in section \ref{sec:pre}. 
In Fig. \ref{fig: tsencode}, for Sensor \RomanNumeralCaps{1} and Sensor \RomanNumeralCaps{2}, we perform rotation shift ($\rho$) twice to $\mathcal H_{t_1}$ and $\mathcal H'_{t_1}$, once to $\mathcal H_{t_2}$ and $\mathcal H'_{t_2}$, and keep $\mathcal H_{t_3}$ and $\mathcal H'_{t_3}$  the same. 
We bind data samples in one sampling widow by calculating $\mathcal H = \rho\rho\mathcal H_{t_1} * \rho\mathcal H_{t_2} * \mathcal H_{t_3}$ and $\mathcal H' = \rho\rho\mathcal H'_{t_1} * \rho\mathcal H'_{t_2} * \mathcal H'_{t_3}$. Finally, to integrate data from multiple sensors, we generate a random signature hypervector for each sensor and bind information as 
$\mathcal S_1 * \mathcal H_1 + \mathcal S_2  * \mathcal H_2 + \ldots + \mathcal S_n * \mathcal H_n, $
where $\mathcal S_i$ denote the signature hypervector for Sensor $i$, and $\mathcal H_i$ denote the data from sensor $i$. 
In Fig. \ref{fig: tsencode}, we combine information from Sensor \RomanNumeralCaps{1} and Sensor \RomanNumeralCaps{2} by randomly generating sigature hypervectors $\mathcal S$ and $\mathcal S'$ and calculating $\mathcal S * \mathcal H + \mathcal S' * \mathcal H'$.

\subsection{Domain-Specific Modeling} \label{sec:domain-specific}
\begin{algorithm}[t]
\small
  \caption{Domain-Specific Modeling}\label{alg:adapt}
  \begin{algorithmic}[1]{}
\Require{$\mathcal N$ encoded training samples $\mathcal H_1, \mathcal H_2, \ldots,  \mathcal H_{\mathcal N}$ with $n$ classes and $k$ domains, domains for training data $\mathcal G_1, \mathcal G_2, \ldots, \mathcal G_k$, label of each data sample $\mathcal L_1, \mathcal L_2, \ldots, \mathcal L_\mathcal N$, learning rate $\eta$. }
    
    \Ensure{$k$ domain-specific models $\mathcal M_1, \mathcal M_2, \ldots ,\mathcal M_{k}$ after one training iteration.}
    
    \ForEach{$\lambda \in [1, k]$}
        \State{Initialize a domain-specific model $\mathcal M_\lambda$ consisting of one class hypervectors for each class $\mathcal M_\lambda = \{\mathcal C_{\lambda}^1, \mathcal C_{\lambda}^2, \ldots, \mathcal C_{\lambda}^n\}$}
        \ForEach {$\mathcal H_i \in \{\mathcal H_1, \mathcal H_2, \ldots,  \mathcal H_{\mathcal N}\}$}
            \If{\textit{domain}$(\mathcal H_i) = \mathcal G_\lambda$}  \label{alg:adapt: domain}
                \State $\mathcal C_{\text{max}} = \arg\max_{\mathcal C_{\lambda}^t}\{\delta(\mathcal H_i, \mathcal C_{\lambda}^1), \ldots, \delta(\mathcal H_i, \mathcal C_{\lambda}^n)\}$  \label{alg:adapt:cos}
                \If{$\mathcal L_i = \mathcal C_{\text{max}}$}
                    \State continue
                \ElsIf{$\mathcal L_i \neq \mathcal C_{\text{max}} \land \mathcal L_i = \mathcal C_{\lambda}^j $} \label{alg:adapt:incorrect}
                    \State $\mathcal C_{\text{max}} \leftarrow \mathcal C_\text{max} -\eta\cdot[1-\delta(\mathcal H_i, \mathcal C_{\text{max}} ) ]\times \mathcal H_i$ \label{alg:adapt: update_begin}
                    \State $\mathcal C_{\lambda}^j \leftarrow \mathcal C_{\lambda}^j +\eta\cdot[1-\delta(\mathcal H_i, \mathcal C_{\lambda}^ j ) ]\times \mathcal H_i$ \label{alg:adapt: update_end}
                \EndIf
            \EndIf
        \EndFor
        \State $\mathcal M_\lambda = \textit{Normalize}(\mathcal M_\lambda)$ \label{alg:adapt:normalize}
    \EndFor
    \State \Return $\mathcal M_1, \mathcal M_2, \ldots ,\mathcal M_{k}$
  \end{algorithmic}
\end{algorithm}

As demonstrated in Fig. \ref{fig: flow}, after mapping training data samples to high-dimensional space $(\encircle{A})$ utilizing the time-series encoding technique in section $\ref{sec:encode}$, $\Design$ separates training data into $k$ subsets ($k$ = the number of domains) based on their domains ($\encircle B$).
We then exploit an efficient and lightweight hyperdimensional learning algorithm to generate a domain-specific model for each domain ($\encircle D$). 
Our approach aims to provide accurate classification performance by identifying common patterns during training and eliminating model saturations. 
As shwon in Algorithm \ref{alg:adapt}, we bundle encoded data points by scaling a proper weight to each of them depending on how much new information is added to class hypervectors. 
For each domain $\mathcal G_\lambda (1\leq \lambda \leq k)$, a new encoded training sample $\mathcal H$ updates the domain-specific model $\mathcal M_\lambda$ based on its cosine similarities with all class hypervectors (line \ref{alg:adapt:cos}), i.e., 
    $$\delta( {\mathcal H}, {\mathcal C_{\lambda}^t})=\frac{ {\mathcal H} \cdot {\mathcal C_{\lambda}^t}}{\| {\mathcal H}\|\cdot \| {\mathcal C_{\lambda}^t}\|}=\frac{ {\mathcal H}}{\| {\mathcal H}\|}\cdot \frac{{\mathcal C_{\lambda}^t}}{\| {\mathcal C_{\lambda}^t}\|}\propto {\mathcal H}\cdot {\textit{Normalize}({\mathcal C_{\lambda}^t})} $$ 
where $ {\mathcal H} \cdot {\mathcal C_{\lambda}^t} (1\leq t\leq n)$ is the dot product between ${\mathcal H}$ and a class hypervector ${\mathcal C_{\lambda}^t}$ representing class $t$ of domain $\mathcal G_\lambda$. 
Here $\| {\mathcal H}\|$ is a constant factor when comparing a query with all classes and thus can be eliminated. 
The cosine similarity calculation can hence be simplified to a dot product operation between $\mathcal H$ and the normalized class hypervector. 
If the prediction matches the expected output, no update will be made to prevent overfitting. 
If ${\mathcal H}$ has the highest cosine similarity with class $\mathcal C_\text{max}$ while its true label matches $\mathcal C_\lambda^j$ (line \ref{alg:adapt:incorrect}), the model updates following lines \ref{alg:adapt: update_begin} - \ref{alg:adapt: update_end}. 
A large $\delta(\mathcal H, \cdot)$ indicates the input data point is marginally mismatched or already exists in the model, and the model is updated by adding a very small portion of the encoded query ($1-\delta(\mathcal H, \cdot) \approx 0$). 
In contrast, a small $\delta(\mathcal H, \cdot)$, indicating a noticeably new pattern that is uncommon or does not already exist in the model, updates the model with a large factor ($1-\delta(\mathcal H, \cdot) \approx 1$). 
We then normalize each  dimension in every model at the end of our algorithm (line \ref{alg:adapt:normalize}). 
Our learning algorithm provides a higher chance for non-common patterns to be properly included in the model, and effectively reduces computationally-expensive retraining iterations required to achieve reasonable accuracy.

\subsection{Domain Generalization}
\begin{algorithm}[t]
\small
  \caption{Domain Generalization}\label{alg:dg}
  \begin{algorithmic}[1]{}
    \Require{$k$ domain-specific models $\{\mathcal M_1, \mathcal M_2, \ldots,  \mathcal M_k\}$ each with size $n \times \mathcal D$, regeneration rate $\mathcal R$. }
    
    \Ensure{Domain-variant dimensions $\mathcal U$ to drop.}
    \State Initialize $n$ empty matrices $\mathcal T_1 , \mathcal T_2, \ldots, \mathcal T_n$, each with size $k\times \mathcal D$. 
    \ForEach{$ \mathcal T_\gamma \in \{\mathcal T_1 , \mathcal T_2, \ldots, \mathcal T_n\}$} \label{alg:dg:cs}
        \ForEach{$ \mathcal M_\lambda \in \{\mathcal M_1, \mathcal M_2, \ldots,  \mathcal M_k\}$}\label{alg:dg:ds}
            \State $\mathcal T_\gamma[\lambda, : ] = \mathcal M_\lambda[\gamma, :]$ \Comment{Form $n$ class-specific matrices} \label{alg:dg:extract}
        \EndFor
        \State$\sigma_\gamma = \textit{Variance}(\mathcal T_\gamma, \textit{columnwise})$  \Comment{\textit{dim}$(\sigma_\gamma)=1\times \mathcal D$} \label{alg:dg:var}
    \EndFor
    \State $\mathcal V = \sum_{i=1}^n \sigma_i$ \label{alg:dg:sum_var}
    \State $\mathcal U = \textrm{argsort}(\mathcal V)\big[\lfloor{(1-\mathcal R)\cdot \mathcal D}\rfloor : \mathcal D\big]$ \Comment{$\mathcal R$ of dimensions with largest variance} \label{alg:dg:select}
    \State \Return $\mathcal U$
  \end{algorithmic}
\end{algorithm}
\setlength{\textfloatsep}{0pt}
In each training iteration, after applying the hyperdimensional learning algorithm, $\Design$ utilizes partially trained models to calculate the relevance of each dimension to domains and filter-out dimensions representing domain-variant information. 
As demonstrated in Fig. \ref{fig: flow}, our \textit{domain generalization} consists of three parts: class-specific aggregation (\encircle{E}), domain-variant filter (\encircle{F}), and model ensemble (\encircle{I}). 
In \textit{class-specific aggregation}, we extract class hypervectors from each domain-specific model obtained in section \ref{sec:domain-specific} to construct class-specific matrices. 
In \textit{domain-variant filter}, $\Design$ identifies and regenerates domain-variant dimensions to enhance the generalization capability of our model. 
In \textit{model ensemble}, we combine multiple domain-specific models into a single model to form the domain-invariant HDC model.  

\begin{figure}[!t]
\centering
\vspace{-3mm}
\includegraphics[width=0.98\linewidth]{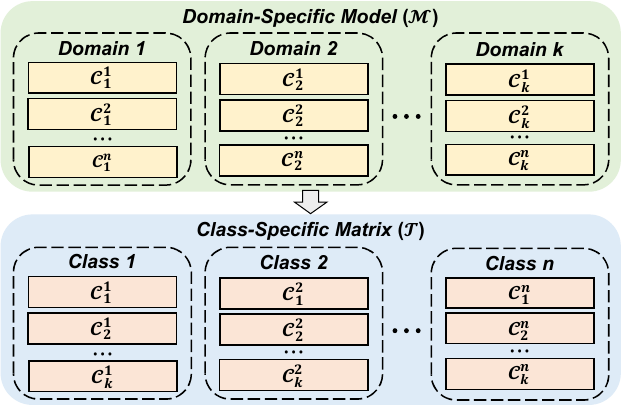}
\vspace{-2mm}
\caption{Extracting Class Hypervectors From $k$ Domain-Specific Models to Contruct $n$ Class-Specific Matricies. Each Class Hypervector is in size $1\times \mathcal D$}
\vspace{1mm}
\label{fig: extraction}  
\end{figure}

\textbf{Class-Specific Aggregation:}
As demonstrated in Algorithm \ref{alg:dg}, for each class, we extract the class hypervector from each domain-specific model representing that class to form a class-specific matrix (line \ref{alg:dg:extract}). 
As shown in Fig. \ref{fig: extraction}, we denote $\mathcal C_{\lambda}^t$ as the class hypervector of class $t$ $(1\leq t \leq n)$ in the domain-specific model $\mathcal M_\lambda$ $(1\leq \lambda \leq k)$, where $k$ and $n$ represent the number of domains and the number of classes, respectively. 
Then, for instance, to construct the class-specific matrix for class 1, we extract $\mathcal C_1^1$ from $\mathcal M_1$, $\mathcal C_2^1$ from $\mathcal M_2$, \ldots, and $\mathcal C_k^1$ from $\mathcal M_k$. Note that we obtain $k$ domain-specific models each with size $n\times\mathcal D$ (Algorithm \ref{alg:dg} line \ref{alg:dg:ds}), where $\mathcal D$ denote the the dimensionality of our HDC models. 
Hence, here we construct $n$ class-specific matrices (line \ref{alg:dg:cs}), each with size $k \times \mathcal D$.

\textbf{Domain-Variant Filter:} 
HDC algorithms represent each class with a hypervector that encodes the patterns of that class. Dimensions with very different values indicate that they store differentiated patterns, while dimensions with similar values indicate they store common information across classes.
Hence, as detailed in Algorithm \ref{alg:dg}, for each class-specific matrix, we calculate the variance of each dimension to measure its dispersion (line \ref{alg:dg:var}). 
Dimensions with large variance indicate that, for the same class, these dimensions store very differentiated information, and are hence considered domain-variant. 
We sum up the variance vector of each class-specific matrix to obtain $\mathcal V$, an $1\times\mathcal D$ vector representing the overall relevance of each dimension to domains (line \ref{alg:dg:sum_var}). 
We then select the top $\mathcal R$ portion of dimensions with the highest variance, where $\mathcal R$ denote the regeneration rate, and filter them out from the model (line \ref{alg:dg:select}). 
To mitigate the performance degradation caused by eliminating dimensions from models, we replace them with new randomly-generated dimensions and retrain them (Fig. \ref{fig: flow} \encircle{G}, \encircle{H}). 
Considering the non-linear relationship between features, we utilize an encoding method inspired by the Radial Basis Function (RBF)\cite{rahimi2007random}. 
Mathematically, for an input vector in original space $\mathcal F = \{f_1, f_2, \ldots, f_n\}(f_i\in \mathds{R}
)$, we generate the corresponding hypervector $\mathcal H=\{h_1, h_2, \ldots, h_{\mathcal{D}}\} (0 \leq h_i \leq 1, h_i \in  \mathds{R})$ with ${\mathcal D} 
 (\mathcal D\gg n)$ dimensions by calculating a dot product of $\mathcal F$ with a randomly generated vector as $h_i = \cos(\mathcal B_i\cdot \mathcal F+c)\times \sin (\mathcal B_i\cdot \mathcal F)$, where $\mathcal B_i = \{b_1, b_2, \ldots, b_n\}$ is a randomly generated base vector with $b_i\sim \textit{Gaussian}(\mu =0,\sigma =1) \textrm{ and } c\sim \textit{Uniform}[0, 2\pi].$ 
$\Design$ replaces each base vector of the selected dimensions in the encoding module with another randomly generated vector from the Gaussian distribution, and retrains the current domain-specific models. 
Instead of training from scratch, $\Design$ only updates values of class hypervectors on the dropped dimensions while other dimensions continue learning based on their existing values. 

\textbf{Model Ensemble:} 
After filtering out and regenerating domain-variant dimensions, we assemble all the domain-specific models to build a general domain-invariant model. 
We construct this model based on domain-specific models and the weight of each domain. 
For each domain $\lambda\in [0,k]$ ($k$ = number of domains), we calculate the proportion of the data from domain $\lambda$ as $\mathcal P_\lambda= \frac{\mathcal N_\lambda}{\mathcal N_{total}}$, where $\mathcal N_\lambda$ denote the number of data samples from domain $\lambda$ and $\mathcal N_{total}$ denote the total number of training samples. 
We then ensemble domain-specific models $\mathcal M_1, \mathcal M_2, \ldots, \mathcal M_k$ into a general domain-invariant model $\mathcal M$ by computing 
$\mathcal M = \mathcal P_1\cdot \mathcal M_1 + \mathcal P_2\cdot \mathcal M_2 + \ldots +\mathcal P_k\cdot \mathcal M_k.$ Note that $\mathcal M$ and each $\mathcal M_\lambda$ are of size $n\times \mathcal D$.

\subsection{Hardware Optimizations}
Edge-based ML applications often involve strict latency and performance requirements. We apply the following hardware-aware optimizations into the implementation of our proposed $\Design$ to maximize its performance and robustness:

\begin{itemize}[leftmargin=*]
    \item \textit{Multithreading.} $\Design$ involves several implicit parallel processing opportunities. We utlizes multi-threads to parallelize operations including random basis generation, basis regeneration, encoding, and vector normalization. 
    \item \textit{Tiled matrix multiplication.} Several steps in  $\Design$ are naturally matrix operations, e.g., encoding and cosine similarity. We tile  matrix operations with memory hierarchy in mind to optimize for the best local memory locality.
    \item \textit{Kernel fusion.} We create custom kernels to fuse original kernels to reduce kernel invocation overheads. 
    \item \textit{Quantization.} $\Design$ supports custom bitwidth to leverage low-bitwidth functional units on hardware platforms to further improve computational efficiency.
\end{itemize}

\section{Evaluation} 
\label{sec:eval}
\begin{table}[!t]
\footnotesize
\caption{Detailed Breakdowns of Datasets($\mathcal N$: number of data samples)}
\vspace{-4.5mm}
\begin{center}
\begin{tabular}{cc|cc|cc} \toprule
\multicolumn{2}{c}{DSADS~\cite{barshan2014recognizing}}& \multicolumn{2}{c}{USC-HAD~\cite{zhang2012usc}} &\multicolumn{2}{c}{PAMAP2~\cite{reiss2012introducing}} \\
\midrule
 Domains & $\mathcal N$  & Domains  & $\mathcal N$ &  Domains  &  $\mathcal N$\\
\midrule
Domain 1 & 2,280 & Domain 1 &  8,945	 &	 Domain 1 & 5,636\\
Domain 2 & 2,280 & Domain 2  &  8,754 &  Domain 2  & 5,591\\
Domain 3 & 2,280  & Domain 3 & 8,534 &  Domain 3 & 5,806\\
Domain 4 & 2,280  & Domain 4 & 8,867  &  Domain 4 & 5,660\\
 & & Domain 5& 8,274& &\\ \midrule
Total & 9,120 & Total & 43,374 & Total &22,693\\
\bottomrule
\end{tabular}
\label{tb:detail_data}
\end{center}
\vspace{-2mm}
\end{table}
Taking human activity recognition as the application use-case, we evaluate $\Design$ on widely-used multi-sensor time series datasets DSADS~\cite{barshan2014recognizing}, USC-HAD~\cite{zhang2012usc}, PAMAP2~\cite{reiss2012introducing}.
Domains are defined by subject grouping chosen based on subject ID from low to high.
The data size of each domain in each dataset is demonstrated in TABLE \ref{tb:detail_data}.
We compare $\Design$ with (\romannumeral 1) two SOTA CNN-based domain generalization (DG) algorithms: Representation Self-Challenging (RSC)~\cite{huang2020self} and AND-mask~\cite{parascandolo2020learning}, and (\romannumeral 2) SOTA HDC without domain generalization capability~\cite{hernandez2021onlinehd} (BaselineHD).
Our evaluations include leave-one-domain-out (LODO) performance, 
learning efficiency on both server CPU and resource-constrained devices,
 and robustness against hardware noises. 
The CNN-based DG algorithms are trained with TensorFlow, and we utilize the common practice of grid search to identify the best hyper-parameters for each model.
The results of BaselineHD are reported in two dimensionality: (\romannumeral 1) \textit{Physical dimensionality} ($\mathcal D=0.5$\textrm{k}) of $\Design$, a compressed dimensionality designed for resource-constrained computing platforms, (\romannumeral 2) \textit{Effective dimensionality} ($\mathcal D^*=4\textrm{k}$), defined as the sum of the physical dimensions $(\mathcal D)$ of $\Design$ with all the regenerated dimensions throughout the retraining iterations. Mathematically, $\mathcal D^*=\mathcal D+\mathcal D \times \mathcal R \times \textit{Number of Iterations}$, where $\mathcal R$ is the regeneration rate.
We also explore the hyperparameter design space of $\Design$ to identify optimal hyperparameters. 
\subsection{Experimental Setup}
To evaluate the performance of $\Design$ on both high-performance computing environments and resource-limited devices, we include results from the following platforms:
\begin{itemize}[leftmargin=*]
  \item \textbf{Server CPU}: Intel Xeon Silver 4310 CPU (12-core, 24-thread, 2.10 GHz), 96 GB DDR4 memory, Ubuntu 20.04, Python 3.8.10, PyTorch 1.12.1, TDP 120 W.
  \item \textbf{Embedded CPU}: Raspberry Pi 3 Model 3+ (quad-core ARM A53 @1.4GHz), 1 GB LPDDR2 memory, Debian 11, Python 3.9.2, PyTorch 1.13.1, TDP 5 W.
  \item \textbf{Embedded GPU}: Jetson Nano (quad-core ARM A57 @1.43 GHz, 128-core Maxwell GPU), 4 GB LPDDR4 memory, Python 3.8.10, PyTorch 1.13.0, CUDA 10.2,  TDP 10 W. 
\end{itemize}


\subsection{Data Preprocessing}
We describe the data processing steps for each dataset primarily focusing on data segmentation and domain labeling.
For specific details, please refer to their respective papers.

\textbf{DSADS}\cite{barshan2014recognizing}: The Daily and Sports Activities Dataset includes 19 activities performed by eight subjects. 
Each data segment is a non-overlapping five-second window sampled at 25Hz.
Four domains are formed with two subjects each.

\textbf{USC-HAD}~\cite{zhang2012usc}: The USC human activity dataset includes 12 activities performed by 14 subjects.
Each data segment is a 1.26-second window sampled at 100Hz with 50\% overlap.
Five domains are formed with three subjects each.

\textbf{PAMAP2}~\cite{reiss2012introducing}: The Physical Activity Monitoring dataset includes 18 activities performed by nine subjects.
Each data segment is a 1.27-second window sampled at 100Hz with 50\% overlap.
Four domains, excluding subject nine, are formed with two subjects each.
We also removed invalid and irrelevant data as suggested by the authors.
Lastly, we only retain common activity labels (1, 2, 3, 4. 12, 13, 16, and 17).

\vspace{-1mm}
\subsection{Accuracy}
\begin{figure}[!t]
\centering
\includegraphics[width=\linewidth]{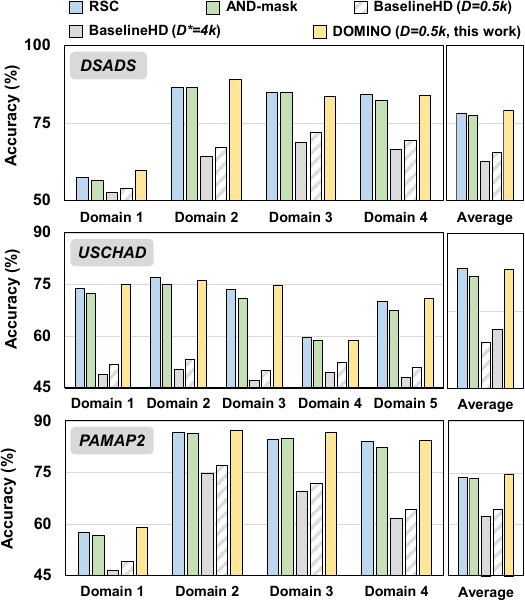}
\vspace{-6mm}
\caption{LODO Accuracy of $\Design$ and CNN-based Approaches}
\label{fig: acc}  
\vspace{2mm}
\end{figure}

The accuracy of the LODO classification is shown in Fig. \ref{fig: acc}. The accuracy of Domain $i$ means that the model is trained  with data from all other domains and tested on data from Domain $i$. 
This accuracy score indicates the generalization capability of a trained model to data from unseen distributions. 
$\Design$ outperforms SOTA CNN-based DG approaches by achieving on average $0.96\%$ higher accuracy than RSC and $2.04\%$ higher accuracy than AND-mask. $\Design$ also exhibits $11.70\%$ higher accuracy than BaselineHD $(\mathcal D^*=4\textrm{k})$ and $15.93\%$ higher accuracy than BaselineHD $(\mathcal D=0.5\textrm{k})$. This shows $\Design$ effectively filters out domain-variant dimensions and thereby improves model generalization capabilities. Additionally, with the informative time series encoding technique and our dynamic regenerating method, $\Design$ delivers reasonable accuracy with notably lower dimensionalities. 

\begin{figure}[!t]
\centering
\includegraphics[width=\linewidth]{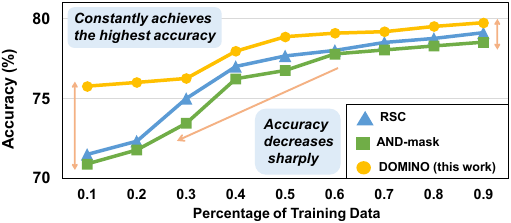}
\vspace{-7.5mm}
\caption{Comparing Performance on Partial Training Data}
\vspace{-4mm}
\label{fig: partial_train}  
\end{figure}

\textbf{Accuracy with Partial Training Data: }
In practical implementation, edge-based ML applications frequently encounter scenarios wherein only a limited portion of data is labeled.
We exhibit the performance of $\Design$ with decreasing amounts of labeled data on the dataset DSADS.
We randomly sample a portion (ranging from $10\%$ to $90\%$) of training data from all the domains except Domain 5 for training, and evaluate our model by using data from Domain 5 for inference. 
As shown in Fig. \ref{fig: partial_train}, with the decreasing of data size, the learning accuracy of SOTA CNN-based DG algorithms decreases sharply while $\Design$ is capable of constantly delivering significantly better performance. 
In particular, when we only use $10\%$ of the training data, $\Design$ demonstrates $5.81\%$ higher accuracy than AND-mask and $4.90\%$ high accuracy than RSC. 

\begin{figure}[!t]
\centering
\includegraphics[width=\linewidth]{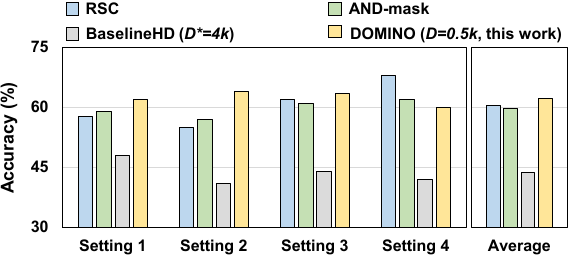}
\vspace{-7mm}
\caption{Comparing Performance on Imbalanced Training Data}
\label{fig:imbalance}  
\end{figure}

\textbf{Accuracy with Imbalanced Training Data:}
Imbalanced training data, where certain domains contribute a disproportionately larger portion of the data while other domains are represented by considerably smaller amounts, often severely degrades model performance. 
We evaluate the performance of $\Design$ when learning from highly imbalanced training data using the dataset USC-HAD as demonstrated in Fig. \ref{fig:imbalance}.
In  Setting $\gamma (1\leq \gamma \leq 4)$, we construct the training data by randomly sampling $70\%$ of the data from Domain $\gamma$ and the remaining $30\%$ from all other domains except Domain 5. 
We evaluate our trained model by using data from Domain 5 for inference. 
$\Design$ outperforms SOTA CNN-based DG approaches by exhibiting 
on average $1.18\%$ and $2.58\%$ higher accuracy than RSC and AND-mask, respectively.


\begin{figure}[!t]
\centering
\includegraphics[width=\linewidth]{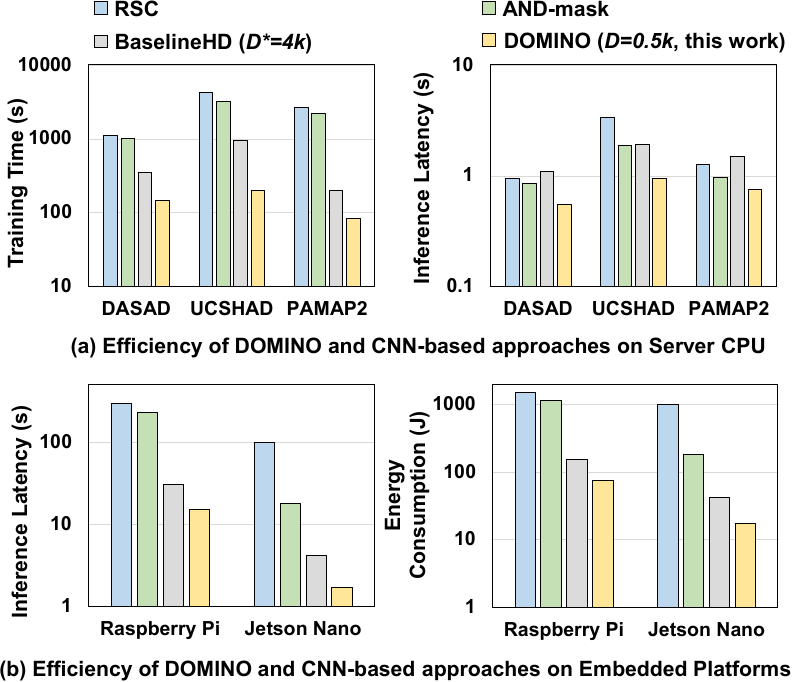}
\vspace{-6mm}
\caption{ \textbf{(log scale)} Comparing Efficiency on Server GPU and Edge Platforms}
\label{fig: efficiency}  
\vspace{3mm}
\end{figure}

\subsection{Efficiency}

\textbf{Efficiency on Server CPU:} 
For fairness, we compare the learning efficiency of $\Design$ with RSC, AND-mask, and BaselineHD ($\mathcal D=4\textrm{k}$) on the server CPU. 
For each dataset, each domain consists of roughly similar amounts of data as detailed in TABLE \ref{tb:detail_data}; thus, we show the average runtime of training and inference for all the domains. 
As demonstrated in Fig. \ref{fig: efficiency}(a), $\Design$ exhibits $16.34\times$ faster training than RSC and $14.17\times$ faster training than AND-mask. 
Additionally, $\Design$ delivers $2.89\times$ faster inference than RSC and $1.97\times$ faster inference than AND-mask. 
Such notable higher learning efficiency is thanks to the highly parallel matrix operations on high-dimensional space and the notably faster convergence of HDC. 
Compared to BaselineHD ($\mathcal D^* = 4$k), $\Design$ delivers considerably higher accuracy without sacrificing much training efficiency as shown in Fig. \ref{fig: acc} and Fig. \ref{fig: efficiency}. 
$\Design$ also achieves $1.93\times$ faster inference than BaselineHD ($\mathcal D^* = 4$k) since it requires considerably lower physical dimensionality ($\mathcal D$) and thereby reduces large amounts of unnecessary computations.

\textbf{Efficiency on Embedded CPU and GPU:} 
To further understand the performance of $\Design$ on resource-constrained platforms, we evaluate the efficiency of $\Design$, RSC, and AND-mask using a Raspberry Pi 3 Model B+ board and an NVIDIA Jetson Nano board. 
Both platforms have very limited memory and CPU cores (and GPU cores for Jetson Nano). 
Fig. \ref{fig: efficiency}(b) shows the average inference latency for each algorithm processing each domain in the DSADS dataset. 
$\Design$ outperforms CNN-based DG algorithms by providing speedups $19.79\times$ compared to RSC and $15.31\times$ compared to AND-mask on Raspberry Pi.
$\Design$ also delivers $58.44\times$ faster inference than RSC and $10.49\times$ faster inference than AND-mask on Jetson Nano. 
Additionally, $\Design$ exhibits significantly less energy consumption, indicating that $\Design$ can run efficiently on energy-constrained platforms. 

\begin{figure}[!t]
\centering
\includegraphics[width=\linewidth]{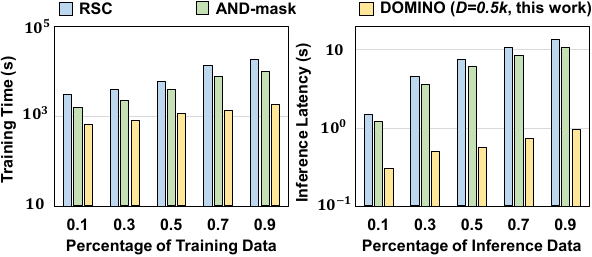}
\vspace{-7mm}
\caption{Comparing Scalability Using Different Size of Data}
\vspace{-3mm}
\label{fig: scalability}  
\end{figure}
\subsection{Data Size Scalability} 
We exhibit the scalability of $\Design$ and SOTA CNN-based DG approaches using various training data sizes (percentages of the full dataset). As shown in Fig. \ref{fig: scalability}, with the increasing size of the training dataset, $\Design$ is capable of maintaining high efficiency in both training and inference with a sub-linear growth in execution time. In contrast, the training and inference time of CNN-based algorithms increases notably faster than $\Design$.  This indicates that $\Design$ is capable of providing timely and scalable DG solutions for both high-performance and resource-constrained computing devices.
\begin{figure}[!t]
\centering
\includegraphics[width=0.9\linewidth]{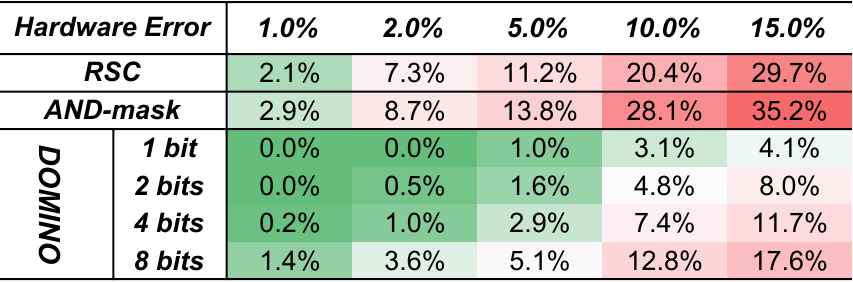}
\vspace{-2mm}
\caption{Comparing Quality Loss Under Hardware Errors}
\label{fig: robust}  
\end{figure}

\vspace{-1mm}
\subsection{Robustness Against Hardware Noises}

Leveraging encoded data points on high-dimensional space, $\Design$ exhibits ultra-robustness against noise and failures, ensuring effective performance of DG classification tasks on noisy embedded devices. 
In particular, each hypervector stores information across all its components and no component is more responsible for storing any more information than another. 
We compare the robustness of $\Design$ with SOTA CNN-based DG algorithms against hardware noise by showing the average quality loss under different percentages of hardware errors in Fig. \ref{fig: robust}. 
The error rate refers to the percentage of random bit flips on memory storing CNNs and $\Design$ models. 
For fairness, all CNNs weights are quantized to their effective 8-bit representation. 
In CNNs, random bit flip results in significant quality loss as corruptions on most significant bits can cause major weight changes. 
In contrast, $\Design$ provides significantly higher robustness against noise due to its redundant and holographic distribution of patterns with high dimensionality. Specifically, all dimensions equally contribute to storing information; thus, failures on partial data will not cause the loss of entire information. 
$\Design$ exhibits the maximum robustness using hypervectors in 1-bit precision, that is on average $10.93\%$ higher robustness than AND-mask and $8.62\%$ higher robustness than RSC. 
Increasing precision lowers the robustness of $\Design$ since random flips on more significant bits will cause more loss of accuracy. 

\subsection{Design Space Exploration} \label{sec:eval:dse} 
\begin{figure}[!t]
\centering
\vspace{-1mm}
\includegraphics[width=\linewidth]{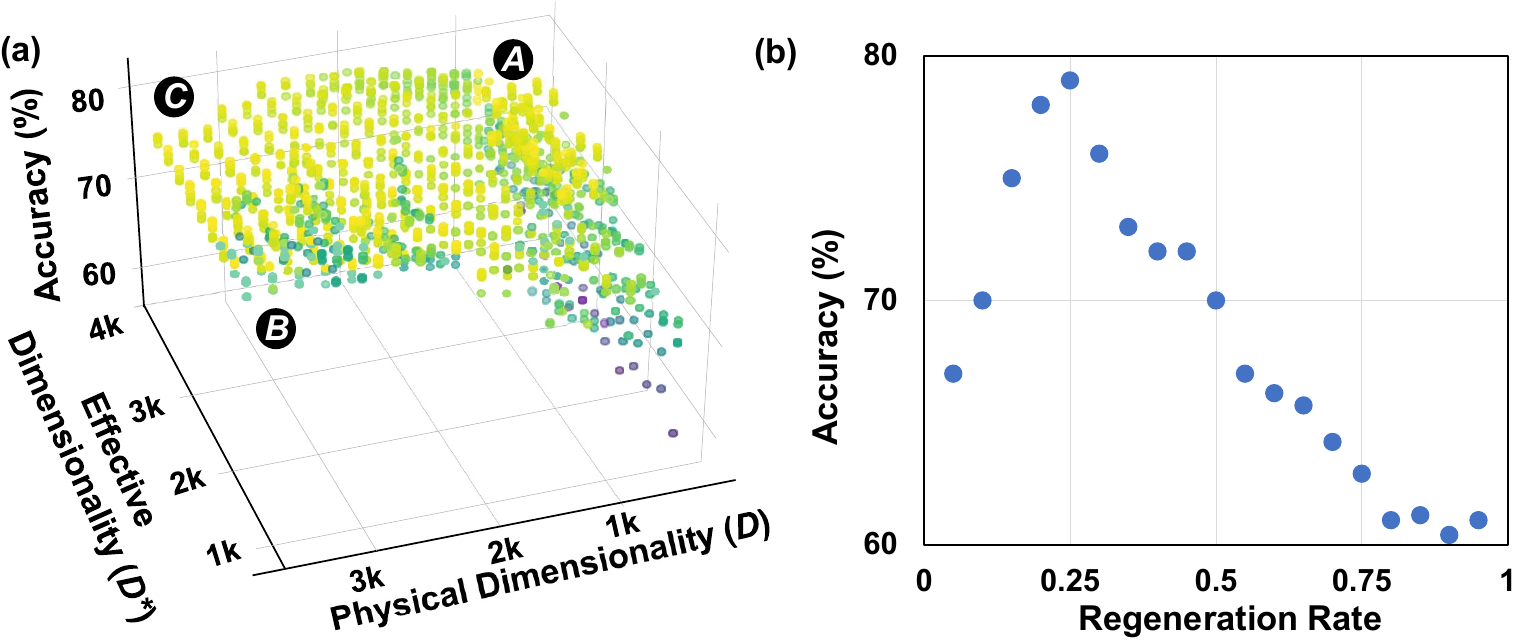}
\vspace{-5mm}
\caption{Relation Between Accuracy and (a) Physical Dimensionality $(\mathcal D)$ and Effective Dimensionality $(\mathcal D^*)$ (b) Regeneration Rate $(\mathcal R)$}
\vspace{2mm}
\label{fig: dse}  
\end{figure}

To understand the behavior of $\Design$ in different hyperparameter settings, 
we conduct a design space exploration traversing all possible combinations of different physical dimensionality $(\mathcal D)$, effective dimensionality $(\mathcal D^*)$, and regeneration rate$(\mathcal R)$. 
We observe that a larger effective dimensionality
$(\mathcal D^*)$ play significant roles in enhancing the model performance. 
For instance, in Fig. \ref{fig: dse}(a), though starting from a much smaller $\mathcal D$, the DG accuracy of points near $\encircle A$ is notably higher than points near $\encircle B$ and is comparable to points near $\encircle C$, as the disadvantage of utilizing a very limited physical dimensionality can be compensated by more training iterations and larger effective dimensionality. 
This characteristics is extremely appealing since using models with compressed dimensionalities significantly reduces unnecessary computations involved in training and inference and can be more resource-efficient for edge-based ML applications. 
Mathematically, $\textit{Number of Iterations} = \frac{\mathcal D^*-\mathcal D}{\mathcal D\times \mathcal R}$. 
This indicates when $\mathcal R$ and $\mathcal D$ are fixed, models with larger $\mathcal D^*$ go through more iterations of filtration and regeneration of domain-variant dimensions and potentially perform better.  
We also learn the relation between accuracy and $\mathcal R$ as shown in Fig. \ref{fig: dse}(b) by fixing $\mathcal D$ and $\mathcal D^*$ to $0.5\textrm{k}$ and $4\textrm{k}$, respectively. We observe $\Design$ achieves optimal performance when $\mathcal R$ is around $0.25$. A too small $\mathcal R$ will cause failure in filtering out the majority of domain-invariant dimensions, while a too large $\mathcal R$ will lead to performance degradation due to excessively eliminating and regenerating informative dimensions. 

\section{Related Works} \label{sec:related}
\subsection{Distribution Shift}
Distribution shift (DS), where a model is deployed on a data distribution different from what it was trained on, poses significant robustness challenges in real-world ML applications~\cite{dong2022first, palakkadavath2022improving, gulrajani2021in}. 
Various innovative ideas have been proposed to mitigate this issue, and can be majorly categorized as \textit{domain generalizations} (DG) and \textit{domain adaptations} (DA).  
DA 
generally utilizes unlabeled or sparsely labeled data in target domains to quickly adapt a model trained in different source domains~\cite{csurka2017domain, li2018learning}. 
In contrast, most DG approaches aim to build models 
by extracting domain-invariant features across known domains~\cite{li2018learning, li2018domain, dou2019domain}. 
Existing DG are primarily based on CNNs~\cite{huang2020self, ganin2016domain, sagawadistributionally, li2018learning, parascandolo2020learning} and target image classification and object detection tasks. 
However, these approaches often rely on multiple convolutional and fully-connected layers, requiring intensive computations and iterative refinement. 
This can often be challenging to implement on resource-constrained platforms considering the memory and power limitations~\cite{pan2017future}. 
Our proposed $\Design$ is the first HDC-based DG algorithm aiming to provide a more resource-efficient and hardware-friendly DG solution for today's edge-based ML applications.  
It utilizes encoded data points on high-dimensional space to dynamically identify and regenerate domain-variant dimensions, thereby enhancing model generalization capability.

\subsection{Hyperdimensional Computing} 
Prior studies have exhibited enormous success in various applications of HDCs, such as brain-like reasoning~\cite{poduval2022graphd}, bio-signal processing~\cite{burrello2019laelaps}, and cyber-security~\cite{wang2023late}. 
A few endeavors have been made toward utilizing HDC for time series classification, including data from Electroencephalography (EEG) and Electromyography (EMG) sensors~\cite{ni2022neurally,moin2021wearable, rahimi2016hyperdimensional}.
However, existing HDCs do not consider the challenge of distribution shift. 
For instance, TempHD~\cite{ni2022neurally} relies on historical EEG data from an individual to make inferences for the same individual. 
This can be a detrimental drawback, especially in the deployment of embedded AI applications. 
Hyperdimensional Feature Fusion~\cite{wilson2023hyperdimensional}, a recently proposed algorithm for out-of-distribution (OOD) sample detection, maps information from multiple layers of DNNs into class-specific vectors in high-dimensional space to deliver ultra-efficient performance. 
However, its proposed algorithm remains to rely on resource-intensive multi-layer DNNs, and a systematic way to deal with OOD samples for DG has not yet been proposed. 
In contrast, we propose $\Design$, an HDC-based DG learning framework that fully leverages the highly efficient and parallel matrix operations on high-dimensional space to filter out dimensions representing domain-variant information. 

\section{Conclusion} \label{sec:conclude}
In this paper, we propose $\Design$, an innovative HDC-based DG algorithm for noisy multi-sensor time series classification. 
Our evaluations demonstrate that $\Design$ achieves notably higher accuracy than CNN-based DG approaches, especially when learning from partially labeled data and highly imbalanced data. 
Additionally, $\Design$ provides a hardware-friendly solution for both high-performance computing devices and resource-constrained platforms by delivering significantly faster training and inference on both server CPU and embedded platforms. 
Leveraging holographic pattern distributions on high-dimensional space, $\Design$ also exhibits considerably higher robustness against hardware errors than CNN-based approaches, bringing unique advantages in performing learning tasks on noisy and unstable edge devices. 
\section{Acknowledgement}
This work was partially supported by the National Science Foundation (NSF) under award CCF-2140154.

\newpage
\bibliographystyle{IEEEtran}
\bibliography{reference}

\begin{thebibliography}{10}
\providecommand{\url}[1]{#1}
\csname url@samestyle\endcsname
\providecommand{\newblock}{\relax}
\providecommand{\bibinfo}[2]{#2}
\providecommand{\BIBentrySTDinterwordspacing}{\spaceskip=0pt\relax}
\providecommand{\BIBentryALTinterwordstretchfactor}{4}
\providecommand{\BIBentryALTinterwordspacing}{\spaceskip=\fontdimen2\font plus
\BIBentryALTinterwordstretchfactor\fontdimen3\font minus
  \fontdimen4\font\relax}
\providecommand{\BIBforeignlanguage}[2]{{%
\expandafter\ifx\csname l@#1\endcsname\relax
\typeout{** WARNING: IEEEtran.bst: No hyphenation pattern has been}%
\typeout{** loaded for the language `#1'. Using the pattern for}%
\typeout{** the default language instead.}%
\else
\language=\csname l@#1\endcsname
\fi
#2}}
\providecommand{\BIBdecl}{\relax}
\BIBdecl

\bibitem{wang2023disthd}
J.~Wang, S.~Huang, and M.~Imani, ``Disthd: A learner-aware dynamic encoding
  method for hyperdimensional classification,'' \emph{arXiv preprint
  arXiv:2304.05503}, 2023.

\bibitem{shrivastwa2020brain}
R.~R. Shrivastwa \emph{et~al.}, ``A brain--computer interface framework based
  on compressive sensing and deep learning,'' \emph{IEEE Consumer Electronics
  Magazine}, 2020.

\bibitem{rashid2022template}
N.~Rashid \emph{et~al.}, ``Template matching based early exit cnn for
  energy-efficient myocardial infarction detection on low-power wearable
  devices,'' \emph{Proceedings of the ACM on Interactive, Mobile, Wearable and
  Ubiquitous Technologies}, 2022.

\bibitem{demirel2022neural}
B.~U. Demirel \emph{et~al.}, ``Neural contextual bandits based dynamic sensor
  selection for low-power body-area networks,'' in \emph{Proceedings of the
  ACM/IEEE International Symposium on Low Power Electronics and Design}, 2022.

\bibitem{dong2022first}
\BIBentryALTinterwordspacing
K.~Dong and T.~Ma, ``First steps toward understanding the extrapolation of
  nonlinear models to unseen domains,'' in \emph{NeurIPS 2022 Workshop on
  Distribution Shifts: Connecting Methods and Applications}, 2022. [Online].
  Available: \url{https://openreview.net/forum?id=lfs4KqfrY1}
\BIBentrySTDinterwordspacing

\bibitem{pooch2020can}
E.~H. Pooch \emph{et~al.}, ``Can we trust deep learning based diagnosis? the
  impact of domain shift in chest radiograph classification,'' in
  \emph{Thoracic Image Analysis: Second International Workshop, TIA 2020, Held
  in Conjunction with MICCAI 2020, Lima, Peru, October 8, 2020, Proceedings
  2}.\hskip 1em plus 0.5em minus 0.4em\relax Springer, 2020.

\bibitem{subbaswamy2020development}
A.~Subbaswamy \emph{et~al.}, ``From development to deployment: dataset shift,
  causality, and shift-stable models in health ai,'' \emph{Biostatistics},
  2020.

\bibitem{wilson2023hyperdimensional}
S.~Wilson \emph{et~al.}, ``Hyperdimensional feature fusion for
  out-of-distribution detection,'' in \emph{Proceedings of the IEEE/CVF Winter
  Conference on Applications of Computer Vision}, 2023.

\bibitem{gulrajani2021in}
\BIBentryALTinterwordspacing
I.~Gulrajani \emph{et~al.}, ``In search of lost domain generalization,'' in
  \emph{International Conference on Learning Representations}, 2021. [Online].
  Available: \url{https://openreview.net/forum?id=lQdXeXDoWtI}
\BIBentrySTDinterwordspacing

\bibitem{wang2018deep}
M.~Wang \emph{et~al.}, ``Deep visual domain adaptation: A survey,''
  \emph{Neurocomputing}, 2018.

\bibitem{zhao2020review}
S.~Zhao \emph{et~al.}, ``A review of single-source deep unsupervised visual
  domain adaptation,'' \emph{Transactions on Neural Networks and Learning
  Systems}, 2020.

\bibitem{wilson2020multi}
G.~Wilson \emph{et~al.}, ``Multi-source deep domain adaptation with weak
  supervision for time-series sensor data,'' in \emph{Proceedings of the 26th
  ACM SIGKDD international conference on knowledge discovery \& data mining},
  2020.

\bibitem{qin2017dual}
Y.~Qin \emph{et~al.}, ``A dual-stage attention-based recurrent neural network
  for time series prediction,'' \emph{arXiv preprint arXiv:1704.02971}, 2017.

\bibitem{su2019robust}
Y.~Su \emph{et~al.}, ``Robust anomaly detection for multivariate time series
  through stochastic recurrent neural network,'' in \emph{Proceedings of the
  25th ACM SIGKDD international conference on knowledge discovery \& data
  mining}, 2019.

\bibitem{hochreiter1997long}
S.~Hochreiter \emph{et~al.}, ``Long short-term memory,'' \emph{Neural
  computation}, 1997.

\bibitem{ni2022neurally}
Y.~Ni \emph{et~al.}, ``Neurally-inspired hyperdimensional classification for
  efficient and robust biosignal processing,'' in \emph{Proceedings of the 41st
  IEEE/ACM International Conference on Computer-Aided Design}, 2022.

\bibitem{imani2019framework}
M.~Imani \emph{et~al.}, ``A framework for collaborative learning in secure
  high-dimensional space,'' in \emph{CLOUD}.\hskip 1em plus 0.5em minus
  0.4em\relax IEEE, 2019.

\bibitem{ge2020classification}
L.~Ge \emph{et~al.}, ``Classification using hyperdimensional computing: A
  review,'' \emph{IEEE Circuits and Systems Magazine}, 2020.

\bibitem{zou2021scalable}
Z.~Zou \emph{et~al.}, ``Scalable edge-based hyperdimensional learning system
  with brain-like neural adaptation,'' in \emph{SC}, 2021.

\bibitem{wang2023late}
J.~Wang, H.~Chen, M.~Issa, S.~Huang, and M.~Imani, ``Late breaking results:
  Scalable and efficient hyperdimensional computing for network intrusion
  detection,'' \emph{arXiv preprint arXiv:2304.06728}, 2023.

\bibitem{rahimi2007random}
A.~Rahimi and B.~Recht, ``Random features for large-scale kernel machines,''
  \emph{Advances in neural information processing systems}, 2007.

\bibitem{barshan2014recognizing}
B.~Barshan \emph{et~al.}, ``Recognizing daily and sports activities in two open
  source machine learning environments using body-worn sensor units,''
  \emph{The Computer Journal}, 2014.

\bibitem{zhang2012usc}
M.~Zhang \emph{et~al.}, ``Usc-had: A daily activity dataset for ubiquitous
  activity recognition using wearable sensors,'' in \emph{Proceedings of the
  2012 ACM conference on ubiquitous computing}, 2012.

\bibitem{reiss2012introducing}
A.~Reiss \emph{et~al.}, ``Introducing a new benchmarked dataset for activity
  monitoring,'' in \emph{16th international symposium on wearable
  computers}.\hskip 1em plus 0.5em minus 0.4em\relax IEEE, 2012.

\bibitem{huang2020self}
Z.~Huang \emph{et~al.}, ``Self-challenging improves cross-domain
  generalization,'' in \emph{Computer Vision--ECCV 2020: 16th European
  Conference, Glasgow, UK, August 23--28, 2020, Proceedings, Part II 16}.\hskip
  1em plus 0.5em minus 0.4em\relax Springer, 2020.

\bibitem{parascandolo2020learning}
G.~Parascandolo \emph{et~al.}, ``Learning explanations that are hard to vary,''
  \emph{arXiv preprint arXiv:2009.00329}, 2020.

\bibitem{hernandez2021onlinehd}
A.~Hernandez-Cane \emph{et~al.}, ``Onlinehd: Robust, efficient, and single-pass
  online learning using hyperdimensional system,'' in \emph{Design, Automation
  \& Test in Europe Conference \& Exhibition (DATE)}.\hskip 1em plus 0.5em
  minus 0.4em\relax IEEE, 2021.

\bibitem{palakkadavath2022improving}
R.~Palakkadavath \emph{et~al.}, ``Improving domain generalization with
  interpolation robustness,'' in \emph{NeurIPS 2022 Workshop on Distribution
  Shifts: Connecting Methods and Applications}, 2022.

\bibitem{csurka2017domain}
G.~Csurka \emph{et~al.}, \emph{Domain adaptation in computer vision
  applications}.\hskip 1em plus 0.5em minus 0.4em\relax Springer, 2017.

\bibitem{li2018learning}
D.~Li \emph{et~al.}, ``Learning to generalize: Meta-learning for domain
  generalization,'' in \emph{Proceedings of the AAAI conference on artificial
  intelligence}, 2018.

\bibitem{li2018domain}
H.~Li \emph{et~al.}, ``Domain generalization with adversarial feature
  learning,'' in \emph{Proceedings of the IEEE conference on computer vision
  and pattern recognition}, 2018.

\bibitem{dou2019domain}
Q.~Dou \emph{et~al.}, ``Domain generalization via model-agnostic learning of
  semantic features,'' \emph{Advances in Neural Information Processing
  Systems}, 2019.

\bibitem{ganin2016domain}
Y.~Ganin \emph{et~al.}, ``Domain-adversarial training of neural networks,''
  \emph{The journal of machine learning research}, 2016.

\bibitem{sagawadistributionally}
S.~Sagawa \emph{et~al.}, ``Distributionally robust neural networks,'' in
  \emph{International Conference on Learning Representations}, 2019.

\bibitem{pan2017future}
J.~Pan \emph{et~al.}, ``Future edge cloud and edge computing for internet of
  things applications,'' \emph{IEEE Internet of Things Journal}, 2017.

\bibitem{poduval2022graphd}
P.~Poduval \emph{et~al.}, ``Graphd: Graph-based hyperdimensional memorization
  for brain-like cognitive learning,'' \emph{Frontiers in Neuroscience}, 2022.

\bibitem{burrello2019laelaps}
A.~Burrello \emph{et~al.}, ``Laelaps: An energy-efficient seizure detection
  algorithm from long-term human ieeg recordings without false alarms,'' in
  \emph{Design, Automation \& Test in Europe Conference \& Exhibition
  (DATE)}.\hskip 1em plus 0.5em minus 0.4em\relax IEEE, 2019.

\bibitem{moin2021wearable}
A.~Moin \emph{et~al.}, ``A wearable biosensing system with in-sensor adaptive
  machine learning for hand gesture recognition,'' \emph{Nature Electronics},
  2021.

\bibitem{rahimi2016hyperdimensional}
A.~Rahimi \emph{et~al.}, ``Hyperdimensional biosignal processing: A case study
  for emg-based hand gesture recognition,'' in \emph{International Conference
  on Rebooting Computing}.\hskip 1em plus 0.5em minus 0.4em\relax IEEE, 2016.

\end{thebibliography}
\end{document}